\documentclass[letterpaper]{article} 
\usepackage{aaai24}  
\usepackage{times}  
\usepackage{helvet}  
\usepackage{courier}  
\usepackage[hyphens]{url}  
\usepackage{graphicx} 
\usepackage{natbib}  
\usepackage{caption} 
\usepackage{booktabs}
\usepackage{algorithm}
\usepackage{algorithmic}
\usepackage{subcaption}
\usepackage{enumitem}
\usepackage{amsmath}
\usepackage{amssymb}
\usepackage{mathtools}
\usepackage{amsthm}
\usepackage{thmtools} 
\usepackage{thm-restate}
\usepackage[T1]{fontenc}
\theoremstyle{plain}
\newtheorem{theorem}{Theorem}[section]
\newtheorem{example}[theorem]{Example}
\newtheorem{proposition}[theorem]{Proposition}

\theoremstyle{definition}
\newtheorem{definition}[theorem]{Definition}

\theoremstyle{remark}

\DeclareMathOperator{\E}{\mathbb{E}}
\DeclareMathOperator{\R}{\mathbb{R}}
\usepackage{newfloat}
\usepackage{listings}
\DeclareCaptionStyle{ruled}{labelfont=normalfont,labelsep=colon,strut=off} 
\floatstyle{ruled}
\newfloat{listing}{tb}{lst}{}
\floatname{listing}{Listing}
\title{Non-linear Welfare-Aware Strategic Learning}
\author{
    Tian Xie\textsuperscript{\rm 1},
    Xueru Zhang\textsuperscript{\rm 1}
}
\affiliations{
    \textsuperscript{\rm 1}Department of Computer Science and Engineering, the Ohio State University\\
    \{xie.1379, zhang.12807\}@osu.edu\\
}

\usepackage{bibentry}

\begin{document}

\maketitle

\begin{abstract}
    This paper studies algorithmic decision-making in the presence of strategic individual behaviors, where an ML model is used to make decisions about human agents and the latter can adapt their behavior strategically to improve their future data. Existing results on strategic learning have largely focused on the \textit{linear} setting where agents with linear labeling functions best respond to a (noisy) linear decision policy. Instead, this work focuses on general non-linear settings where agents respond to the decision policy with only "local information" of the policy. Moreover, we simultaneously consider objectives of maximizing \textit{decision-maker welfare} (model prediction accuracy), \textit{social welfare} (agent improvement caused by strategic behaviors), and \textit{agent welfare} (the extent that ML underestimates the agents). We first generalize the agent best response model in previous works to the non-linear setting and then investigate the compatibility of welfare objectives. We show the three welfare can attain the optimum simultaneously only under restrictive conditions which are challenging to achieve in non-linear settings. The theoretical results imply that existing works solely maximizing the welfare of a subset of parties usually diminish the welfare of others. We thus claim the necessity of balancing the welfare of each party in non-linear settings and propose an \textit{irreducible} optimization algorithm suitable for general strategic learning. Experiments on synthetic and real data validate the proposed algorithm. 
\end{abstract}

\section{Introduction}\label{sec:intro}

When machine learning (ML) models are used to automate decisions about human agents (e.g., in lending, hiring, and college admissions), they often have the power to affect agent behavior and reshape the population distribution. With the (partial) information of the decision policy,  agents can change their data strategically to receive favorable outcomes, e.g., they may exert efforts to genuinely improve their labels or game the ML models by manipulating their features without changing the labels. This phenomenon has motivated an emerging line of work on \textit{Strategic Classification} \citep{Hardt2016a}, which explicitly takes the agents' strategic behaviors into account when learning ML models. 

One challenge in strategic classification is to model the interactions between agents and the ML decision-maker, as well as understand the impacts they each have on the other. As the inputs of the decision system, agents inevitably affect the utility of the decision-maker. Meanwhile, the deployed decisions can further influence agents' future data and induce societal impacts. Thus, to facilitate socially responsible algorithmic decision-making, we need to consider the welfare of all parties when learning the decision system, including the welfare of the decision-maker, agent, and society. Specifically, the decision-maker typically cares more about assigning correct labels to agents when making decisions, while society will be better off if the decision policy can incentivize agents to improve their actual states (labels). For the agents, it is unrealistic to assign each of them a positive decision, but it is reasonable for agents to expect a "just" decision that does not underestimate their actual states. 

However, most existing works on strategic classification solely consider the welfare of one party. For example, \citet{Dong2018,Braverman2020,Hardt2021,Sundaram2021,zhang2022,Eilat2022} only focus on \textit{decision-maker welfare} by designing accurate classifiers robust to strategic manipulation. \citet{Kleinberg2020,harris2021stateful,ahmadi2022} solely consider maximizing \textit{social welfare} by designing an optimal policy to incentivize the maximum improvement of agents. To the best of our knowledge, only a few works consider both \textit{decision-maker welfare} and \textit{social welfare} \citep{bechavod2022information,Barsotti2022}. Among them, \citet{bechavod2022information} initiate a theoretical study to understand the relationship between the two welfare, where they identify a sufficient condition under which the \textit{decision-maker welfare} and \textit{social welfare} can be optimized at the same time. This paper still focuses on incentivizing agent improvement to maximize \textit{social welfare}. 

Moreover, most prior works limit the scope of analysis to linear settings by assuming both the agent labeling function and the ML decision rules are linear \cite{Dong2018, chen2020learning, ahmadi2021strategic, horowitz2023causal}. This is unrealistic in real applications when the relationship between features and labels is highly intricate and non-linear. Meanwhile, previous works typically modeled strategic behaviors as "(noisy) best responses" to the decision policy \cite{Hardt2016a, Hardt2021,Dong2018, chen2020learning, ahmadi2021strategic, horowitz2023causal}, which is not practical when the decision policy is highly complex and agents' knowledge about the policy (e.g., a neural network) is far from complete. It turns out that the existing theoretical results and methods proposed under linear settings where agents "best respond" can easily be violated in non-linear settings. Thus, there remain gaps and challenges in developing a comprehensive framework for strategic learning under general non-linear settings.

To tackle the limitations in prior works, we study the welfare of all parties in strategic learning under general non-linear settings where both the ground-truth labeling function and the decision policies are complex and possibly non-linear. Importantly, we develop a generalized agent response model that allows agents to respond to policy only based on local information/knowledge they have about the policy. Under the proposed response model, we simultaneously consider: 1) \textit{decision-maker welfare} that measures the prediction accuracy of ML policy; 2) \textit{social welfare} that indicates the level of agent improvement; 3) \textit{agent welfare} that quantifies the extent to which the policy under-estimates the agent qualifications. 

Under general non-linear settings and the proposed agent response model, we first revisit the theoretical results in \citet{bechavod2022information} and show that those results can easily be violated under more complex settings. We then explore the relationships between the objectives of maximizing decision-maker, social, and agent welfare. The results show that the welfare of different parties can only be aligned under rigorous conditions depending on the ground-truth labeling function, agents' knowledge about the decision policy, and the realizability (i.e., whether the hypothesis class of the decision policy includes the ground-truth labeling function). It implies that existing works that solely maximize the welfare of a subset of parties may inevitably diminish the welfare of the others. We thus propose an algorithm to balance the welfare of all parties.

The rest of the paper is organized as follows. After reviewing the related work in Sec.~\ref{sec:related}, we formulate a general strategic learning setting in Sec.~\ref{sec:problem}, where the agent response and the welfare of the decision-maker, agent, and society are presented. In Sec.~\ref{sec:welfare}, we discuss how the existing theoretical results derived under linear settings are violated under non-linear settings and provide a comprehensive analysis of the decision-maker/agent/social welfare; we show all the conditions when they are guaranteed to be compatible in Thm. \ref{thm:general}. This demonstrates the necessity of an "irreducible" optimization framework balancing the welfare of all parties, which we introduce in Sec.~\ref{sec:opt}. Finally, in Sec.~\ref{sec:experiment}, we provide comprehensive experimental results to compare the proposed method with other benchmark algorithms on both synthetic and real datasets.

\section{Related Work}\label{sec:related}

Next, we briefly review the related work and discuss the differences with this paper.

\subsection{Strategic Learning} \citet{Hardt2016a} was the first to formulate strategic classification as a Stackelberg game where the decision-maker publishes a policy first following which agents best respond. A large line of subsequent research has focused on designing accurate classifiers robust to strategic behaviors. For example, \citet{Ben2017,Chen2020,tang2021} designed robust linear regression predictors for strategic classification, while \citet{Dong2018, ahmadi2021strategic, lechner2023strategic, shao2024strategic} studied online learning algorithms for the problem. \citet{levanon2021strategic} designed a protocol for the decision maker to learn agent best response when the decision policy is non-linear, but it assumes agents are perfectly aware of the complex classifiers deployed on them. \citet{Braverman2020,Hardt2021} relaxed the assumption that the best responses of agents are deterministic and studied how the decision policy is affected by randomness. \citet{Sundaram2021} provided theoretical results on PAC learning under the strategic setting, while \citet{LevanonR22,horowitz2023causal} proposed a new loss function and algorithms for linear strategic classification. \citet{Miller2020, shavit2020, Alon2020} focused on causal strategic learning where features are divided into "causal" and "non-causal" ones. However, all these works study strategic classification under linear settings and the primary goal is to maximize \textit{decision-maker welfare}. In contrast, our work considers the welfare of not only the decision-maker but also the agent and society. Another line of work related to strategic learning is \textit{Performative Prediction} \cite{perdomo2020, jin2024addressing} to capture general model-dependent distribution shifts. Despite the generality, the framework is highly abstract and hard to interpret when considering the welfare of different parties. We refer to \citet{hardt2023performative} as a comprehensive survey.

\subsection{Promote Social Welfare in Strategic Learning} Although the definition of social welfare is sometimes subtle \cite{florio2014applied}, \textit{social welfare} is often regarded as the improvement of ground truth qualifications of agents in strategic learning settings and we need to consider how decision rules impact \textit{social welfare} by shaping agent best response. A rich line of works are proposed to design mechanisms to incentivize agents to improve (e.g., \citet{Lydia2019,shavit2020,Alon2020,zhang2020,Chen2020,Kleinberg2020,ahmadi2022,ahmadi2022setting,Jin2022,xie2024automating}). The first prominent work was done by \citet{Kleinberg2020}, where they studied the conditions under which a linear mechanism exists to incentivize agents to improve their features instead of manipulating them. Specifically, they assume each agent has an effort budget and can choose to invest any amount of effort on either improvement or manipulation, and the mechanism aims to incentivize agents to invest all available efforts in improvement. \citet{harris2021stateful} designed a more advanced stateful mechanism considering the accumulative effects of improvement behaviors. They stated that improvement can accumulate over time and benefit agents in the long term, while manipulation cannot accumulate state to state. On the contrary, \citet{estornell2021incentivizing} considered using an audit mechanism to disincentivize manipulation which instead incentivize improvement. Besides mechanism design, \citet{Rose2020} studied incentivizing improvement in machine learning algorithms by adding a look-ahead regularization term to favor improvement. \citet{Chen2020} divided the features into immutable, improvable, and manipulable features and explored linear classifiers that can prevent manipulation and encourage improvement. \citet{Jin2022} also focused on incentivizing improvement and proposed a subsidy mechanism to induce improvement actions and improve social well-being metrics. \citet{bechavod2021} demonstrated the ability of strategic decision-makers to distinguish features influencing the label of individuals under an online setting. \citet{ahmadi2022} proposed a linear model where strategic manipulation and improvement are both present. \citet{Barsotti2022} conducted several empirical experiments when improvement and manipulation are possible and both actions incur a linear deterministic cost. 

The works closest to ours are \citet{bechavod2022information,Rose2020}. \citet{bechavod2022information} provided analytical results of maximizing total improvement without harming anyone's current qualification, but they assumed both the ground truth and the decision rule are linear. \citet{Rose2020} maximized accuracy while punishing the classifier only if it brings the negative externality in non-linear settings. However, solely punishing negative externality is far from ideal. Our work goes further by analyzing the relationships between the welfare of different parties and providing an optimization algorithm to balance the welfare. 

\subsection{Welfare-Aware Machine Learning} The early work on welfare-aware machine learning typically considers balancing \textit{social welfare} and another competing objective such as fairness or profit. For example, \citet{ensign2018runaway} focused on long-term \textit{social welfare} when myopically maximizing efficiency in criminal inspection practice creates a runaway feedback loop. \citet{liu2020disparate,zhang2020,guldogan2022equal} showed that instant fairness constraints may harm \textit{social welfare} in the long run since the agent population may react to the classifier unexpectedly. \citet{rolf2020balancing} proposed a framework to analyze the Pareto-optimal policy balancing the prediction profit and a possibly competing welfare metric but assumed a simplified setting where the welfare of admitting any agent is constant which is not the case in strategic learning. 
\section{Problem Formulation}\label{sec:problem}

Consider a population of strategic agents with features $X = (X_1,...,X_d)$, $\text{dom}(X) \subset \R^d$, and label $Y \in \{0,1\}$ that indicates agent qualification for certain tasks (with ``0" being unqualified and ``1" qualified). Denote $P_X$ as the probability density function of $X$, and assume labeling function $h(x):=P(Y = 1 | X = x)$ is continuous. The agents are subject to certain algorithmic decisions $D(x): \mathbb{R}^d \to \{0,1\}$ (with ``1" being positive and ``0" negative), which are made by a decision-maker using a scoring function $f(x):=P(D(x)=1|X=x)$. Unlike many studies in strategic classification that assume $f \in \mathcal{F}$ is a linear function\footnote{A linear scoring function can be $\mathbf{1}(w^Tx + b \ge 0)$ (e.g., \citet{LevanonR22,horowitz2023causal}) or $w^Tx + b$ when $\text{dom}(X)$ is bounded (e.g., \citet{bechavod2022information}).}, we allow $\mathcal{F}$ to be non-linear. Moreover, it is possible that hypothesis class $\mathcal{F}$ does not include $h$. 

\subsection{Information level and generalized agent response} 

We consider practical settings where the decisions deployed on agents have downstream effects and can reshape the agent's future data. We also focus on "benign" agents whose future labels change according to the fixed labeling function $h(x)$ \cite{raab2021unintended,Rose2020,guldogan2022equal}. Given a function $f$, we define agent best response to $f$ in Def. \ref{def:br_f}, which is consistent with previous works on strategic learning \cite{Hardt2016a,Dong2018,ahmadi2021strategic}.
\begin{definition}[Agent best response to a function]\label{def:br_f}
    Assume an agent with feature $x$ will incur cost $c(x, x')$ by moving her feature from $x$ to $x'$. Then we say the agent best responds to function $f$ if she changes her features from $x$ to $x^{*} = \arg\max_{x'}\{f(x') - c(x,x')\}$.
\end{definition}
It is worth noting that most existing works in strategic learning assume agents either best respond to the exact $f$ (e.g., \citet{Hardt2016a,horowitz2023causal,Dong2018,ahmadi2021strategic}) or noisy $f$ with additive noise (e.g., \citet{LevanonR22, Hardt2021}). However, the best response model in Def.~\ref{def:br_f} requires the agents to know the exact values of $f(x')$ for all $x'$ in the feature domain, which is a rather strong requirement and unlikely holds in practice, especially when policy $f\in \mathcal{F}$ is complex (e.g., logistic regression/neural networks). Instead, it may be more reasonable to assume agents only have "local information" about the policy $f$, i.e., agents with features $x$ only have some information about $f$ at $x$ and they know how to improve their qualifications based on their current standings \cite{zhang2022, raab2021unintended}. To formalize this, we define the agent's \textit{information level} and the resulting agent response in Def.~\ref{def:br}.

\begin{definition}[Information level and agent response]\label{def:br}
Agents with features $x$ and \textit{information level} $K$ know the gradient of $f$ at $x = [x_1,...,x_d]$ up to the $K^{th}$ order (the agents know $\{\nabla f(x), \nabla^2 f(x),..., \nabla^K f(x)\}$ if all these gradients exist). Let $Q_x^f$ be the Taylor expansion of $f$ at $x$ up to the $K^{th}$ order. 
When $f$ is only $k < K$ times differentiable, $Q_x^f$ simply uses all existing gradient information. Then the agents with information level $K$ will respond to $f$ by first estimating $f$ using $Q_x^f$, and then best responding to  $Q_x^f$. 

\end{definition}

Def.~\ref{def:br} specifies the \textit{information level} of agents and how they respond to the decision policy using all the information available at the locality. A lower level of $K$ is more interpretable in practice. For example, agents with information level $K = 1$ use the first-order gradient of $f$ to calculate a linear approximation, which means they only know the most effective direction to improve their current feature values. In contrast, agents with information level $K = 2$ also know how this effective direction to improve will change (i.e., the second-order gradient that reveals the curvature of $f$).

It is worth noting that the agent response under $K=1$ is also similar to \citet{Rose2020} where agents respond to the decision policy by moving toward the direction of the local first-order gradient, and \citet{dean2023emergent} where the agents aim to reduce their risk based on the available information at the current state.  Moreover, Def.~\ref{def:br} can also be regarded as a generalization of the previous models where agents best respond to the exact decision policy under the linear setting. To illustrate this, assume the agents have $K\geq 1$ and the linear policy is $f(x) = \beta^T x$, then the agents with feature $x$ will estimate $f$ using $Q^f_x(x') = f(x) + \nabla f(x) \cdot (x' - x) = \beta^T x'$, which is the same as $f$. However, for complex $f$, agents with information level $K$ may no longer best respond to $f$. Prop. \ref{prop:info} illustrates the necessary and sufficient conditions for an agent to best respond to $f$.


\begin{proposition}\label{prop:info}
    An agent with information level $K$ can best respond to $f$ if and only if $f$ is a polynomial with an order smaller than or equal to $K$.
\end{proposition}

Prop. \ref{prop:info} shows that agents with only local information of policy can only best respond to $f$ that is polynomial in feature vector $X$. In practical applications, $f$ often does not belong to the polynomial function class (e.g., a neural network with \textit{Sigmoid} activation may have infinite order gradients, and the one with \textit{Relu} activation may be a piece-wise function), making it unlikely for agents to best respond to $f$. Note that although Def.~\ref{def:br} specifies the agents to best respond to $Q_x^f$ which is an approximation of actual policy $f$ using the "local information" at $x$, we also present a protocol in Alg.~\ref{alg:response} to learn a more general response function $x^{*} = \Delta_{\phi}(x, I_k(f,x))$ from population dynamics where $\phi$ are learnable parameters and $I_k(f,x)$ is all information available for agents with information level $K$ and feature $x$. This protocol can learn a rich set of responses when the agents have information level $K$ and use their information to respond to $f$ strategically.

\subsection{Welfare of different parties}\label{subsec:welfare}

In this paper, we jointly consider the welfare of the decision-maker, agents, and society. Before exploring their relations, we first introduce the definition of each below.

\paragraph{\textbf{Decision-maker welfare.}} It is defined as the utility the decision maker immediately receives from the deployed decisions, without considering the downstream effects and agent response. Formally, for agents with data $(x,y)$, we define the utility the decision-maker receives from making the decision $D(x)$ as $u(D(x),y)$. Then the \textit{decision-maker welfare} is just the expected utility over the population, i.e., $$\text{DW}(f) = \E_{x,y \sim P_{XY}}[u(D(x),y)].$$  
where $P_{XY}$ is the joint distribution of $(X,Y)$. When $u(D(x),y)=\textbf{1}(D(x)=y)$, $\text{DW}(f)$ reduces to the prediction accuracy. In Sec.~\ref{sec:opt} and \ref{sec:experiment}, we focus on this special case, but the results and analysis can be extended to other utility functions.

\paragraph{\textbf{Social welfare.}} It evaluates the impact of the agent response on population qualifications. We consider two types of social welfare commonly studied in strategic learning: \textit{agent improvement} and \textit{agent safety}. Specifically, agent improvement is defined as:
$$\text{IMP}(f) = \E_{x \sim P_{X}}[h(x^{*}) - h(x)].$$
$\text{IMP}(f)$ measures the increase in the overall population qualification caused by the agent response. This notion has been theoretically and empirically studied under linear settings (e.g., \citet{Kleinberg2020,bechavod2022information}).

However, maximizing improvement should be based on ensuring safety. For agent safety, we define it as: 
$$\text{SF}(f) = \E_{x \sim P_X}\big[\min\big\{h(x^{*}) - h(x),0\big\}\big].$$
It illustrates whether there are agents suffering from deteriorating qualifications after they best respond. This notion has been studied in \citet{bechavod2022information} and was interpreted as "do no harm" constraint, i.e., the deployed decisions should avoid adversely impacting agents. \citet{Rose2020} has shown that \textit{agent safety} can not be automatically guaranteed even under linear settings, and they proposed an algorithm to ensure safety in general settings. However, they did not consider agent improvement which we believe is an equally important measure of social welfare.

\paragraph{\textbf{Agent welfare.}} Previous literature rarely discussed the welfare of agents. The closest notion was \textit{social burden} \citep{milli2019social} defined as the minimum cost a positively-labeled user must pay to obtain a positive decision, where the authors assume agents can only ``game" the decision-maker without improving their true qualifications. Since the agents can always strive to improve in our setting, their efforts are not necessarily a ``burden". However, a relatively accurate policy may intentionally underestimate the current qualification to incentivize agent improvement. Thus, we propose \textit{agent welfare} which measures the extent to which the agent qualification is under-estimated by the decision-maker, i.e., 
$$\text{AW}(f) = \E_{x \sim P_{X}}\big[\min\big\{f(x) - h(x),0\big\}\big].$$
When the policy underestimates the agent's true potential ($f(x)<h(x)$), the agent may be treated unfairly by the deployed decision (i.e., when a qualified agent gets a negative decision). We thus use $\text{AW}(f)$ to measure agent welfare.



\section{Welfare Analysis}\label{sec:welfare}

This section explores the relationships between the decision-maker, social, and agent welfare introduced in Sec.~\ref{sec:problem}. Among all prior works, \cite{bechavod2022information} provides a theoretical analysis of \textit{social} and \textit{decision-maker welfare} under linear settings. In this section, we first revisit the results in \citep{bechavod2022information} and show that they may not be applicable to non-linear settings. Then, we examine the relationships between three types of welfare generally.

\subsection{Beyond the linear setting}\label{subsec:linear}

\citet{bechavod2022information} considered linear settings when both the labeling function $h$ and decision policy $f$ are linear and revealed that \textit{decision-maker welfare} ($\text{DW}(f)$) and \textit{agent improvement} ($\text{IMP}(f)$) can attain the optimum simultaneously when the decision maker deploys $f = h$ and agents best respond to $f$ with a quadratic cost. However, under general non-linear settings, the compatibility of welfare is rarely achieved and we present the necessary and sufficient conditions to guarantee the alignment of all welfare in Thm. \ref{thm:general}.  

\begin{theorem}[Alignment of all welfare]\label{thm:general}
    Suppose agents have information level $K$ and $c(x,x')= (x-x')^TQ(x-x')$ where $Q$ is positive definite, then \textit{decision-maker welfare}, \textit{social welfare}, and \textit{agent welfare} attain maximum simultaneously regardless of the feature distribution $P_X$ and $Q$ if and only if: (i) $h = \hat{w}^{T}x \in \mathcal{F} = \{x \rightarrow w^Tx | w^TRw = c^2\}$ where $R \in \mathbb R^{d\times d}$ is symmetric and positive definite and $c \in \mathbb{R}^{+}$ is a constant; (ii) $\hat{w}$ must be an eigenvector of $R^{-1}Q^{-1}$; (iii) $K \ge 1$.
\end{theorem}

Thm. \ref{thm:general} can be proved based on Prop. \ref{prop:info} together with considerations on maximizing $\text{IMP}(f)$. Intuitively, \textit{decision-maker welfare} and \textit{agent welfare} are guaranteed to be maximized when $f$ is accurate, while safety is only ensured under an accurate $f$ if agents have a perfect understanding of $f$. Moreover, to ensure $\text{IMP}(f)$ is maximized when $f = h$, the function class $\mathcal{F}$ must be restricted to be an ellipsoidal surface. When Thm. \ref{thm:general} is not satisfied, the welfare of different parties is incompatible under the non-linear settings; we use two examples to illustrate this below. 

\begin{example}[Quadratic labeling function with linear policy]\label{example:quadlinear}
    \begin{figure}[h]
        \centering
        \includegraphics[width=0.3\textwidth]{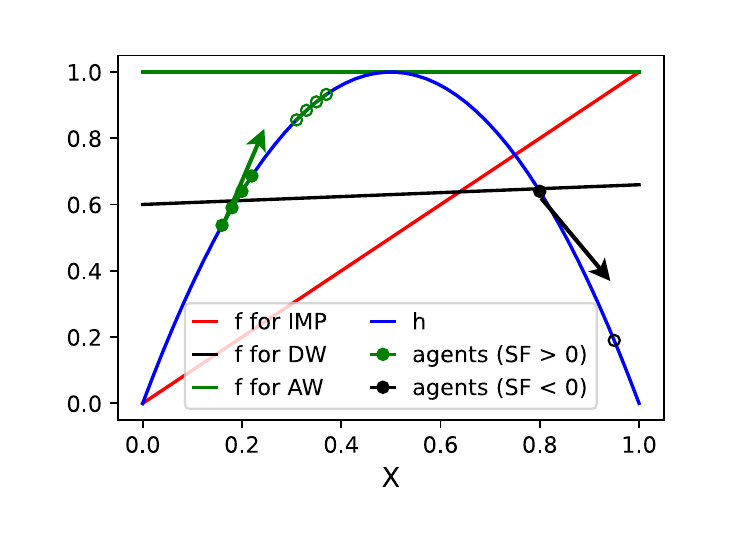}
        \vspace{-0.4cm}
        \caption{Illustration of the optimal linear policies and agent response when $h$ (the blue curve) is quadratic and $f$ is linear. 
        There are four agents with features below $0.5$ (green solid circle) and one agent with the feature above $0.5$ (black solid circle). We observe that (i) the red line maximizes agent improvement $\text{IMP}(f)$ but it is \textbf{unsafe} by only encouraging the green agents to improve while the black agent's qualification deteriorates. The green/black arrows and the hollow circles show how agent response improves/worsens their qualifications; (ii) the black line aims to maximize $\text{DW}(f)$ by running a least square regression with respect to ground truth $h$, which is different from the red line; (iii) Neither the red nor the black line maximizes the \textit{agent welfare} $\text{AW}(f)$; only predictors above $h(x)$ (e.g., constant predictor $f(x) = 1$ shown by green line) achieves the optimal $\text{AW}(f)$.}
        \label{fig:example_quadlinear}
    \end{figure}
    Consider a hiring scenario where a company needs to assess each applicant's fitness for a position. Each applicant has a feature $X \in [0,1]$ demonstrating his/her ability (e.g., coding ability for a junior software engineer position). Suppose the labeling function $h(x) = -4x\cdot(x-1)$ is quadratic with maximum taken at $x = 0.5$, which implies that an applicant's fitness will decrease if the ability is too low ("underqualified") or too high ("overqualified")\footnote{"Overqualified" means an applicant has skills significantly exceeding the requirement. He/she may fit more senior positions or desire a much higher salary than this position can offer.}. Under the quadratic labeling function, if the decision-maker can only select a decision policy $f$ from a linear hypothesis class $\mathcal{F}$ (i.e., violating the requirement that $h \in \mathcal{F}$ in Thm. \ref{thm:general}), then the optimal policies maximizing decision-maker welfare and agent improvement are completely different, i.e., $\text{DW}(f)$ and $\text{IMP}(f)$ cannot be maximized simultaneously. Moreover, agent safety $\text{SF}(f)$ is also not guaranteed. To illustrate this, consider the example in Fig.~\ref{fig:example_quadlinear} with five agents: four green agents with features below $0.5$ and a black one above $0.5$, then the policy that maximizes $\text{IMP}(f)$ can cause the qualification of black agent to deteriorate and lead to negative agent safety $\text{SF}(f)<0$. 
\end{example}

\begin{example}[Quadratic labeling function and policy]\label{example:quadquad}
    Under the same setting of Example \ref{example:quadlinear}, instead of restricting $f$ to be linear (i.e., violating Thm. \ref{thm:general}), let the decision-maker deploy $f(x) = h(x) = -4x \cdot (x-1)$. Obviously, $f$ maximizes both the \textit{decision-maker welfare} and \textit{agent welfare}. However, it can hurt social welfare. Consider an example shown in Fig.~\ref{fig:example_quadquad} where all agents have the same feature $x=0.4$ and information level $K = 1$. Their cost function is $c(x, x') = (x'-x)^2$. Since $\nabla f(x) = -8x + 4 = 0.8$, agents derive $Q^f_x(x') = f(x) + 0.8 \cdot (x' - x) =  0.8x' + 0.64$. Then agent post-response feature will be $x^{*} = x + \frac{0.8}{2} = 0.8$. As a result, the agent qualification becomes $h(x^{*}) = h(0.8) = 0.64 < h(0.4) = 0.96$, which is worse than the initial qualification, i.e., agent safety is not optimal by deploying $f = h$.  
          \begin{figure}[h]
        \centering
\includegraphics[width=0.3\textwidth]{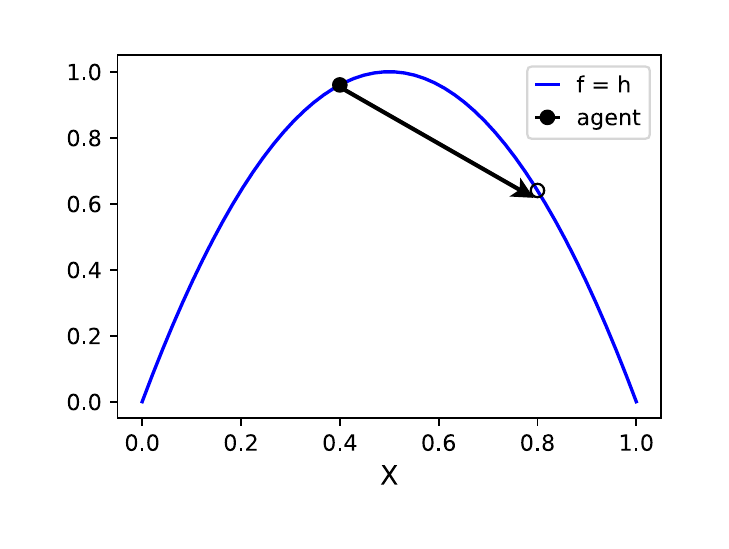}
\vspace{-0.4cm}
        \caption{Violation of agent safety $\text{SF}(f)$ when $f = h$ is quadratic (the blue curve): All agents have the same feature value concentrating on $X = 0.4$ (the solid black circle). If applying the ground truth as the decision policy, all agents will respond by moving to $x^{*} = 0.8$ (the hollow black circle). Although the policy maximizes $\text{AW}(f)$ and $\text{DW}(f)$, all agents become less qualified after they respond, harming the \textit{social welfare}. Both $\text{IMP}(f)$ and $\text{SF}(f)$ are smaller than 0.}
        \label{fig:example_quadquad}
    \end{figure}
\end{example}

Examples \ref{example:quadlinear} and \ref{example:quadquad} illustrate the potential conflict between decision-maker, social, and agent welfare and verify Thm. \ref{thm:general}. Even for simple quadratic settings, the results under the linear setting in \citet{bechavod2022information} no longer hold, and simultaneously maximizing three types of welfare can be impossible. In the following sections, we further analyze the relationships between each pair of welfare in general non-linear settings. 

\subsection{Compatibility analysis of each welfare pair}\label{subsec:general}

Sec.~\ref{subsec:linear} provides the necessary and sufficient condition for the alignment of all welfare. A natural question is whether there exist scenarios under which different pairs of welfare are compatible and can be simultaneously maximized. This section answers this question and explores the relationships between the welfare of every two parties in general settings. 

\subsubsection{Decision-maker welfare \& agent welfare}\label{subsec:da}

We first examine the relationships between \textit{decision-maker welfare} $\text{DW}(f)$ and \textit{agent welfare} $\text{AW}(f)$, the following result shows that the two can be maximized under the same $f$. 

\begin{proposition}
 Under the realizability assumption that $h\in \mathcal{F}$, \textit{decision-maker welfare} and \textit{agent welfare} are compatible and can attain the optimum simultaneously.
\end{proposition}
 Note that \textit{agent welfare} $\text{AW}(f) = 0$ as long as $f(x) \ge h(x)$ holds for all $x$. When $h\in \mathcal{F}$, the decision-maker can simply deploy $f = h$ to assign each agent the correct outcome; this ensures optimal welfare for both parties. 

\subsubsection{Decision-maker welfare \& social welfare}

Unlike compatibility results in Sec~\ref{subsec:da}, the \textit{decision-maker welfare} and either element of \textit{social welfare} (i.e., agent improvement $\text{IMP}(f)$ or agent safety $\text{SF}(f)$) in general are not compatible, and their compatibility needs all conditions in Thm.~\ref{thm:general} to be satisfied.

\begin{proposition}\label{prop:prop2}
Suppose agents have information level $K$ and  $c(x,x')= (x-x')^TQ(x-x')$ where $Q$ is positive definite. The decision-maker welfare and social welfare attain the optimum simultaneously regardless of the feature distribution $P_X$ and $Q$ if and only if Thm.~\ref{thm:general} holds.    
\end{proposition}

As illustrated in Example~\ref{example:quadlinear}, each pair of welfare may be incompatible when $h \notin \mathcal{F}$. Intuitively, since the decision-maker cannot deploy $f = h$, it cannot make accurate decisions and provide a correct direction for agents to improve. Similarly, Example \ref{example:quadquad} shows the incompatibility when $\mathcal{F}$ is non-linear. Although the decision-maker can deploy $f = h$ to attain the optimal $\text{DW}$, the agents may fail to improve because their limited information level produces incorrect estimations of $f$. When $\mathcal{F}$ is non-linear, the gradient $\nabla f(x)$ is no longer a constant, and agents with $K = 1$ will respond in the wrong directions.

\subsubsection{Agent improvement \& agent safety}

While maximum social welfare means maximizing agent improvement $\text{IMP}(f)$ conditioned on perfect agent safety $\text{SF}(f)$, the two in general are not compatible (as shown in Example \ref{example:quadquad}). However, we can identify the necessary and sufficient conditions under which $\text{IMP}(f)$ and $\text{SF}(f)$ attain the maximum simultaneously in general non-linear settings.  

\begin{theorem}[Maximize agent improvement while ensuring safety]\label{theorem:generalswf}
    Suppose agents have information level $K$ and respond to $f$ based on Def. \ref{def:br}. Let $f^{*} \in \arg\max_{f \in \mathcal{F}} \text{IMP(f)}$ be any policy that maximizes the agent improvement, and let $Q^{f^{*}}_x$ be the estimation of $f^{*}$ for agents with feature vector $x$. Then any $f^{*}$ is guaranteed to be safe (i.e., $\text{SF}(f^{*}) = 0$) regardless of the feature distribution $P_X$ and the cost function $c(x,x')$ if and only if: for each agent with feature vector $x$, we have: $\frac{\partial h}{\partial \widetilde{x}_i} \cdot \frac{\partial Q^{f^{*}}_x}{\partial \widetilde{x}_i} \ge 0$ for any $\widetilde{x} = [\widetilde{x}_1,..., \widetilde{x}_d] \in \text{dom}(X)$.
\end{theorem}

This condition implies that for any arbitrary agent with feature $x$, the $f^{*}$ will result in an estimation $Q^{f^{*}}_x$, under which each agent changes her features in directions to improve the qualification.

\subsubsection{Social welfare \& agent welfare}

Finally, we discuss the relationship between \textit{social welfare} and \textit{agent welfare}. Since \textit{agent welfare} is optimal only when $f(x) \ge h(x)$ always holds, $f$ satisfying such a constraint generally cannot maximize \textit{social welfare}. In Example~\ref{example:quadlinear}, a policy that maximizes agent improvement may underestimate certain agents' qualifications and lead to $\text{AW}(f)<0$. Nonetheless, we can still find the necessary and sufficient conditions under which  $\text{IMP}(f)$ and $\text{SF}(f)$ are compatible, as presented below.

\begin{theorem}[Maximize social welfare and agent welfare]\label{theorem:generalaw}
    Suppose agents have information level $K$ and $c(x,x')= (x-x')^TQ(x-x')$ where $Q$ is positive definite. They respond to $f$ based on Def. \ref{def:br}. Denote $\mathcal{F}'=\{f: f \in \mathcal{F} ~and~ f(x) \ge h(x), \forall x \in \text{dom}(X)\}$ as the set of policies maximizing the agent welfare. Then for any $f^{*} \in \arg\max_{f \in \mathcal{F}}\text{IMP}(f) + \text{SF}(f)$, $f^{*}$ is guaranteed to also maximize $\text{AW}(f)$ regardless of the feature distribution $P_X$ and $Q$ if and only if \textbf{either} of the following scenarios holds: (i) Thm. \ref{thm:general} holds; (ii) $\exists f' \in \mathcal{F}'$ and constant $C \in \R^{+}$ such that $Q^{f'}_x = Q^{f^{*}}_x + C$ for any $x \in \text{dom}(X)$.
\end{theorem}

Scenario \textit{(i)} in Thm. \ref{theorem:generalaw} is a sufficient condition, while scenario \textit{(ii)} means that the decision-maker can find a classifier that induces perfect \textit{agent welfare} while simultaneously giving the agents the same information as $f^{*}$. Note that $Q^{f'}_x = Q^{f^{*}}_x + C$ not necessarily implies $f' = f^{*} + C$ when $\text{dom}(X)$ is a zero-measure set. For example, when $K = 2$ and $\text{dom}(X) = \{0\}$, we can have $f^{*} = \text{e}^{x} - x - 1$ and $f' = \frac{x^2}{2} + 0.5$, where $f^{*}$ belongs to the exponential function family but $f'$ belongs to the quadratic function family. 

To conclude, only \textit{decision-maker welfare} and \textit{agent welfare} can be maximized at the same time relatively easily, other pairs of welfare need strong conditions to align under the general settings. Even the two notions in \textit{social welfare} (i.e., agent improvement and safety) can easily contradict each other, and the \textit{decision-maker welfare} and \textit{social welfare} need the most restrictive conditions (which is exactly Thm. \ref{thm:general}) to be aligned. These results highlight the difficulties of balancing the welfare of different parties in general non-linear settings, which further motivates the "irreducible" optimization framework we will introduce in Sec.~\ref{sec:opt}.

\section{Welfare-Aware Optimization}\label{sec:opt}

\begin{algorithm}[h]
\small
    \caption{Strategic Welfare-Aware Optimization (\texttt{STWF})}
    \label{alg:welfare}
    \textbf{Input}: Training set $\mathcal{S}$, Agent response function $\Delta_{\phi}$, Agent information level $K$ \\
    \textbf{Parameters}: Total number of epochs $n$, batch size $B$, learning rate $\gamma$, hyperparameters $\lambda_1, \lambda_2$, initial parameters $\theta$ for $f$\\
    \textbf{Output}: Classifier $f \in \mathcal{F}$
  \begin{algorithmic}[1]
  \STATE Train ground truth model $h \in \mathcal{H}$ with $\mathcal{S}$
    \FOR{$t \in \{1,\ldots n\}$}
         \FOR{$k \in \{1,\ldots \frac{n}{B}\}$}
            \STATE Calculate $\widehat{Y}_B = f_{\theta}(X_B)$
            \STATE Calculate the post-response features of agents: $X_B^{*} = \Delta_{\phi}\bigl(X_B, I_k(f,x))$
            \STATE Calculate the outcomes after responses: $Y_B^{*} = h(X_B^{*})$
            \STATE Calculate $\mathbf{\ell_{DW}}, \mathbf{\ell_{SWF}}, \mathbf{\ell_{AW}}$
            \STATE Compute $L = \mathbf{\ell_{DW}} + \lambda_1 \cdot \mathbf{\ell_{SWF}} + \lambda_2 \cdot \mathbf{\ell_{AW}}$
            \STATE Update $\theta = \theta - \gamma \cdot \frac{\partial L}{\partial \theta}$
    \ENDFOR
    \ENDFOR
    \STATE Return $\theta$
  \end{algorithmic}
\normalsize
\end{algorithm}
 
In this section, we present our algorithm that balances the welfare of all parties in general settings. 
We formulate this welfare-aware learning problem under the strategic setting as a regularized optimization where each welfare violation is represented by a loss term. Formally, we write the optimization problem as follows.
\begin{align}\label{eq:opt}
    \min_{f \in \mathcal{F}} \bigl\{ \mathbf{\ell_{DW}} + \lambda_1 \cdot \mathbf{\ell_{SWF}} + \lambda_2 \cdot \mathbf{\ell_{AW}} \bigr\}
\end{align}
where $\mathbf{\ell_{DW}}$, $\mathbf{\ell_{SWF}}$, $\mathbf{\ell_{AW}}$ are the losses corresponding to \textit{decision-maker welfare}, \textit{social welfare}, and \textit{agent welfare}, respectively. The hyper-parameters $\lambda_1,\lambda_2 \geq 0$ balance the trade-off between the welfare of three parties.  Given training dataset $\{(x_i, y_i)\}_{i=1}^{N}$, the decision-maker first learns $h$ and then calculates each loss as follows:
\begin{itemize}[leftmargin=0.5cm]
  \item $\mathbf{\ell_{DW}}$: we may use common loss functions such as cross-entropy loss, 0-1 loss, etc., and quantify $\mathbf{\ell_{DW}}$ as the average loss over the dataset. In experiments (Sec.~\ref{sec:experiment}), we adopt $\mathbf{\ell_{DW}} = \frac{1}{N} \sum_{i=1}^{N} -\bigl ( y_i  \log\big(f(x_i)\big) + (1-y_i)  \log \big(1 - f(x_i)\big)\bigr).$
  \item $\mathbf{\ell_{SWF}}$: since both \textit{agent improvement} and \textit{agent safety} indicate the \textit{social welfare}. We define $\mathbf{\ell_{SWF}} = \mathbf{\ell_{IMP}} + \mathbf{\ell_{SF}}$, where $\mathbf{\ell_{IMP}}$ is the loss associated with agent improvement and $\mathbf{\ell_{SF}}$ measures the violation of agent safety. In experiments (Sec.~\ref{sec:experiment}), we also present empirical results when $\mathbf{\ell_{SWF}}$ only includes one of the two losses. Specifically, the two losses are defined as follows. 
  \begin{itemize}
    \item $\mathbf{\ell_{IMP}}$: it aims to penalize $f$ when it induces less agent improvement. We can write it as the cross entropy loss over $h(x_i^{*})$ (i.e., the qualification after agent response) and $1$ (i.e., perfectly qualified). Formally, $\mathbf{\ell_{IMP}} = \frac{1}{N} \sum_{i=1}^{N} -\log(h(x_i^*))$. 
    \item $\mathbf{\ell_{SF}}$: it penalizes the model whenever the agent qualification deteriorates after the response. Formally, $\mathbf{\ell_{SF}} = \frac{1}{N} \sum_{x \in \mathcal{A}_{SF}} -\log (h(x^{*}))$ where the set $\mathcal{A}_{SF} = \{x_i | h(x_i^{*}) < h(x_i)\}$ includes all agents with deteriorated qualifications.
  \end{itemize}
  \item $\mathbf{\ell_{AW}}$: it focuses on \textit{agent welfare} and penalizes $f$ whenever the model underestimates the actual qualification of the agent. Formally, $\mathbf{\ell_{AW}} = \frac{1}{N} \sum_{x \in \mathcal{A}_{AW}} -\log (f(x))$ where the set $\mathcal{A}_{AW} = \{x_i | f(x_i) < h(x_i)\}$ includes all agents that are underestimated by the model.
\end{itemize}
The optimizer of ~\eqref{eq:opt} is Pareto optimal and it can be solved using gradient-based algorithms, as long as the gradient of losses for the model parameter (denoted as $\theta$) exists. 
Since $f, h$ are continuous, gradients already exist for $\mathbf{\ell_{DW}}$ and $\mathbf{\ell_{AW}}$. For $\mathbf{\ell_{SWF}}$,  because it is a function of post-response feature $X^{*}$, we need gradients to exist for $X^{*}$ as well. If we already know how agents respond to the model (i.e., $X^{*}$ as a function of $f$ illustrated in Def.~\ref{def:br}), then we only need the $K+1^{th}$ gradients to exist for $f(x)$ with respect to $\theta$. When $K=1$, the objective~\eqref{eq:opt} can be easily optimized if $\nabla f(x)$ is continuous in $\theta$, which is a relatively mild requirement and has also been used in previous literature \citep{Rose2020}. The complete algorithm for \underline{\textbf{S}}tra\underline{\textbf{T}}egic \underline{\textbf{W}}el\underline{\textbf{F}}are-Aware optimization (\texttt{STWF})
is shown in Alg.~\ref{alg:welfare}.

\paragraph{\textbf{Learning agent response function \pmb{$\Delta_\phi$}.}}
Note that while Sec.~\ref{sec:problem} introduces agent response following Def.~\ref{def:br}, our algorithm does not have this requirement; In practice, agent behavior can be more complicated, and we can consider a general learnable response function $x^*=\Delta_{\phi}(x,I_k(f,x))$. Note that \citet{levanon2021strategic} proposed a concave optimization protocol to learn a response function linear in any representation space of agents' features. However, their design implicitly assumes the agents have the perfect knowledge of the representation mapping function used by the decision-maker so that they respond based on their representations, which is unrealistic. Here, we avoid this by only assuming the response is a parameterized function of agents' features $x$ and the information up to level $K$, i.e., $x^{*} = \Delta_{\phi}(x, I_k(f,x))$ where $\phi$ are parameters for some models such as neural networks and $I_k(f,x)$ is all available information for agents with information level $K$ and feature $x$. With this assumption, we provide Alg.~\ref{alg:response} as a general protocol to learn $\phi$. Then we can plug the learned $\Delta_{\phi}$ into \texttt{STWF} (Alg.~\ref{alg:welfare}) to optimize the welfare. 

\begin{algorithm}[h]
\small
    \caption{AgentResponse}
    \label{alg:response}
    \textbf{Input}: Experimental training set $\mathcal{S}$ containing $n$ samples $\{x_j, y_j\}_{j=1}^{n}$,  $N$ arbitrary classifiers $\{f_i\}_{i=1}^{N}$, ground-truth model $h$, agent information level $K$ \\
    \textbf{Parameters}: initial parameters $\phi$ for the agent response function $\Delta_{\phi}(x, I_k(f,x))$ \\
    \textbf{Output}: Parameters $\phi$
  \begin{algorithmic}[1]
    \STATE $\mathcal{S}_0 \leftarrow \{\}$
    \FOR{$i \in \{1,\ldots N\}$}
        \STATE Impose $f_i$ on $\mathcal{S}$ and obtain $x^{*}$ where $x^*$ is the post-response feature vector
        \STATE $\mathcal{S}_i \leftarrow \{\{x_j, I_k(f_i, x_j), x^{*}\}\} $ \COMMENT{build up the new training set for agent response}
        \STATE $\mathcal{S} = \mathcal{S}_{i-1} \cup \mathcal{S}_i$
    \ENDFOR
    \STATE Train $\Delta_{\phi}$ with $\mathcal{S}$
    \STATE Return $\phi$
  \end{algorithmic}
\normalsize
\end{algorithm}

Alg.~\ref{alg:response} requires a population for controlled experiments. As pointed out by \citet{Miller2020}, we can never learn the agent response without the controlled trials. Though the availability of controlled experiments has been assumed in most literature on \textit{performative prediction} \cite{perdomo2020,Izzo2021}, it may be costly to conduct the experiments in practice.
\section{Experiments}\label{sec:experiment}

\begin{table*}[t]
\caption{Comparisons of welfare between \texttt{STWF} and other benchmark algorithms where the largest values are boldfaced.}
\label{results-table:comparison_main}
\setlength\tabcolsep{8pt}
\begin{center}
\resizebox{0.7\textwidth}{!}{
\begin{sc}
\begin{tabular}{p{2.3cm}|l|cccccc}

\toprule
& & \multicolumn{4}{c}{Algorithms} \\
Dataset  & Metric & ERM &  STWF & SAFE & EI & BE \\
\midrule
Synthetic & 
\begin{tabular}{@{}c@{}} Total \\DW  \\ SWF \\ AW  \end{tabular} & 
\begin{tabular}{@{}c@{}}$0.76\pm0.08$\\$0.71\pm0.05$ \\ $0.09\pm0.05$ \\ $-0.22\pm0.001$\end{tabular} & 
\begin{tabular}{@{}c@{}}$\mathbf{1.08}\pm.0.16$\\$\mathbf{0.75}\pm0.08$\\ $\mathbf{0.3}\pm0.08$ \\ $\mathbf{-0.14}\pm0.001$\end{tabular} & 
\begin{tabular}{@{}c@{}}$0.74\pm0.09$\\$0.71\pm0.05$\\ $0.07\pm0.06$ \\ $-0.22\pm0.001$\end{tabular} & 
\begin{tabular}{@{}c@{}}$0.76\pm0.08$\\$0.71\pm0.06$\\ $0.09\pm0.05$ \\ $-0.23\pm0.001$ \end{tabular} &
\begin{tabular}{@{}c@{}}$0.75\pm0.10$\\$0.71\pm0.05$\\ $0.08\pm0.07$\\ $-0.23\pm0.001$ \end{tabular}\\
\midrule
German Credit& 
\begin{tabular}{@{}c@{}} Total \\DW  \\ SWF \\ AW  \end{tabular} & 
\begin{tabular}{@{}c@{}}$\mathbf{0.710}\pm0.10$\\$\mathbf{0.781}\pm0.10$ \\ $0.030\pm0.03$ \\ $-0.100\pm0.00$\end{tabular} & 
\begin{tabular}{@{}c@{}}$0.704\pm.0.14$\\$0.736\pm0.13$\\ $0.022\pm0.04$ \\ $\mathbf{-0.052}\pm0.00$\end{tabular} & 
\begin{tabular}{@{}c@{}}$0.703\pm0.11$\\$0.779\pm0.10$\\ $0.030\pm0.04$ \\ $-0.099\pm0.00$\end{tabular} & 
\begin{tabular}{@{}c@{}}$0.707\pm0.11$\\$0.780\pm0.10$\\ $\mathbf{0.031}\pm0.04$ \\ $-0.100\pm0.00$ \end{tabular} &
\begin{tabular}{@{}c@{}}$0.706\pm0.11$\\$0.777\pm0.10$\\ $0.030\pm0.04$\\ $\-0.100\pm0.00$ \end{tabular}\\
\midrule
ACSIncome-CA& 
\begin{tabular}{@{}c@{}} Total \\DW  \\ SWF \\ AW  \end{tabular} & 
\begin{tabular}{@{}c@{}}$0.74\pm0.06$\\$0.76\pm0.03$ \\ $0.12\pm0.05$ \\ $-0.16\pm0.00$\end{tabular} & 
\begin{tabular}{@{}c@{}}$\mathbf{0.95}\pm.0.15$\\$\mathbf{0.77}\pm0.05$\\ $\mathbf{0.40}\pm0.10$ \\ $-0.21\pm0.00$\end{tabular} & 
\begin{tabular}{@{}c@{}}$0.74\pm0.04$\\$0.76\pm0.03$\\ $0.12\pm0.05$ \\ $\mathbf{-0.15}\pm0.00$\end{tabular} & 
\begin{tabular}{@{}c@{}}$0.74\pm0.06$\\$0.76\pm0.13$\\ $0.12\pm0.04$ \\ $-0.16\pm0.00$ \end{tabular} &
\begin{tabular}{@{}c@{}}$0.73\pm0.08$\\$0.76\pm0.05$\\ $0.12\pm0.05$\\ $-0.16\pm0.00$ \end{tabular}\\
\bottomrule
\end{tabular}
\end{sc}}
\end{center}
\end{table*}

\begin{figure*}[h]
\centering
\begin{subfigure}{0.7\textwidth}
\centering
\includegraphics[width=0.3\textwidth]{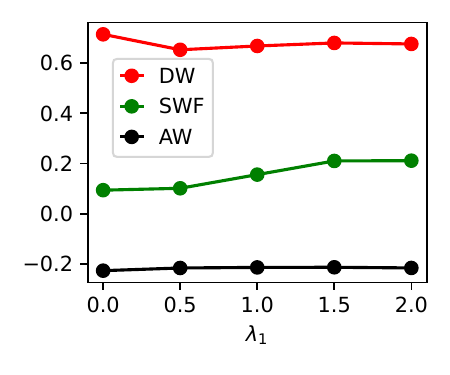}
\includegraphics[width=0.3\textwidth]{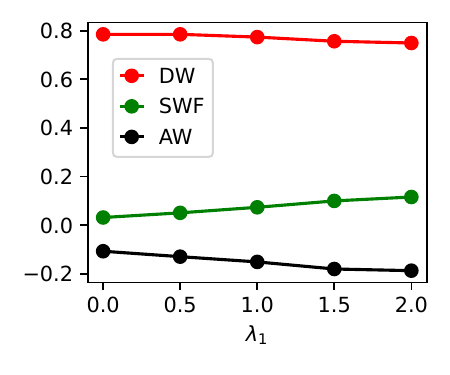}
\includegraphics[width=0.3\textwidth]{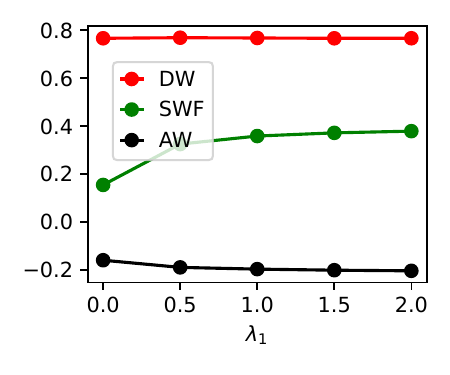}
\vspace{-0.2cm}
\caption{Welfare while increasing $\lambda_1$ and keeping $\lambda_2 = 0$.}
\label{subfig:lambda1}
\end{subfigure}
\begin{subfigure}{0.7\textwidth}
\centering
\includegraphics[width=0.3\textwidth]{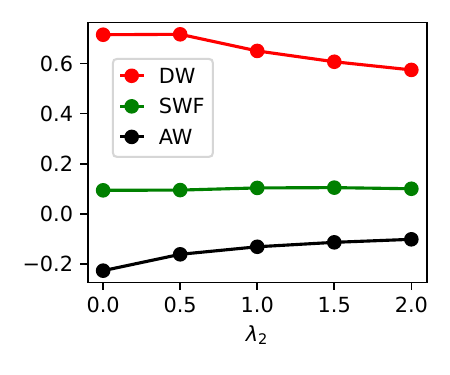}
\includegraphics[width=0.3\textwidth]{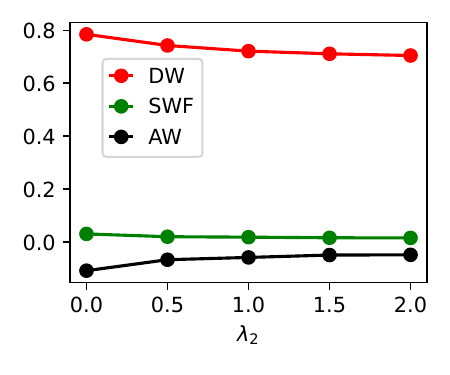}
\includegraphics[width=0.3\textwidth]{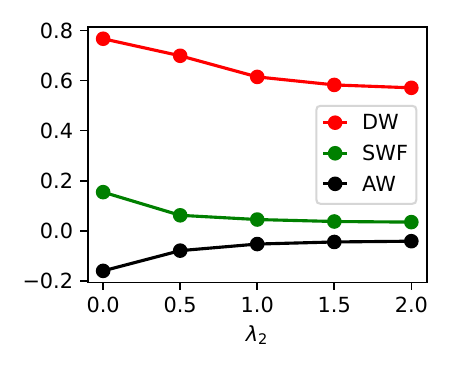}
\vspace{-0.2cm}
\caption{Welfare while increasing $\lambda_2$ and keeping $\lambda_1 = 0$.}
\label{subfig:lambda2}
\end{subfigure}
\caption{Effects of adjusting $\lambda_1, \lambda_2$ on the welfare of each party: synthetic data (left plot), German Credit data (middle plot), ACSIncome-CA data (right plot). Increasing $\lambda_1$ ($\lambda_2$) results in improved $\text{SWF}$ ($\text{AW}$) and trade-offs exist.}
\label{fig:adjust_single}
\end{figure*}

\begin{figure*}[h]
\centering
\begin{subfigure}{0.85\textwidth}
\centering
\includegraphics[width=0.28\textwidth]{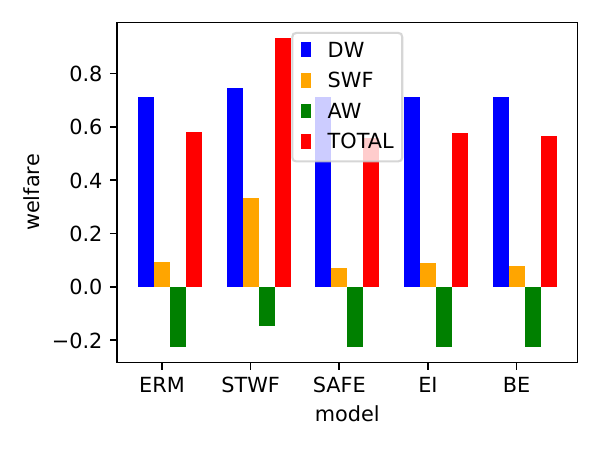}
\includegraphics[width=0.28\textwidth]{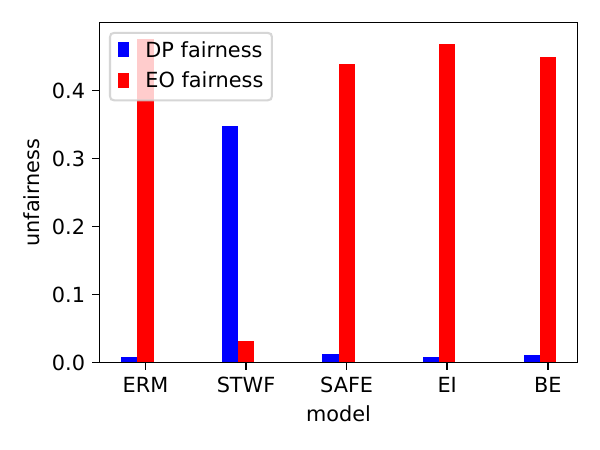}
\includegraphics[width=0.28\textwidth]{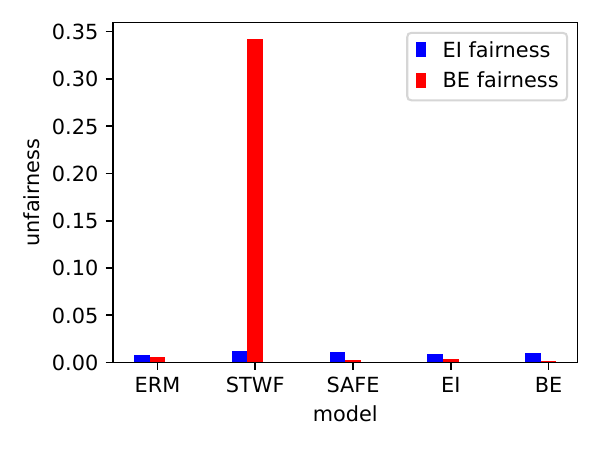}
\caption{Comparisons under the synthetic dataset.}
\label{subfig:quad_comparison}
\end{subfigure}
\hfill
\begin{subfigure}{0.85\textwidth}
\centering
\includegraphics[width=0.28\textwidth]{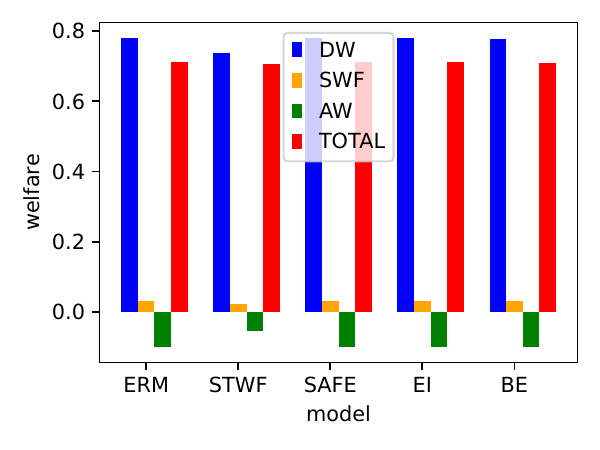}
\includegraphics[width=0.28\textwidth]{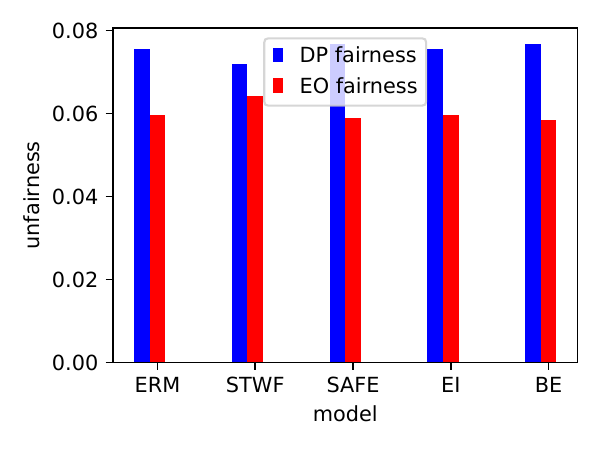}
\includegraphics[width=0.28\textwidth]{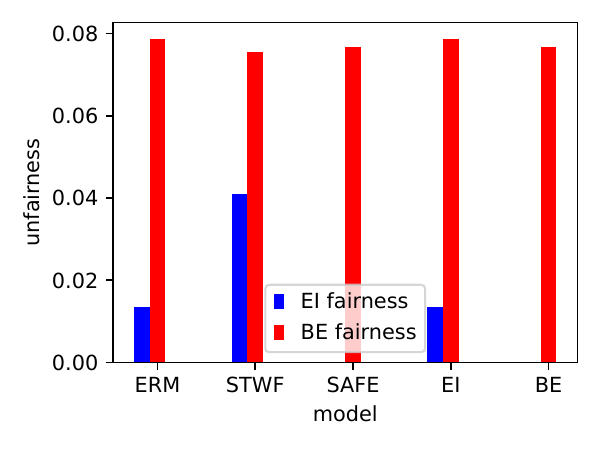}
\caption{Comparisons under the German Credit dataset.}
\label{subfig:german_comparison}
\end{subfigure}
\hfill
\begin{subfigure}{0.85\textwidth}
\centering
\includegraphics[width=0.28\textwidth]{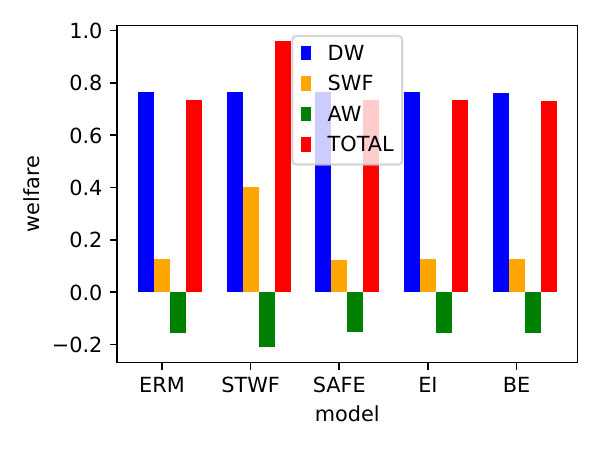}
\includegraphics[width=0.28\textwidth]{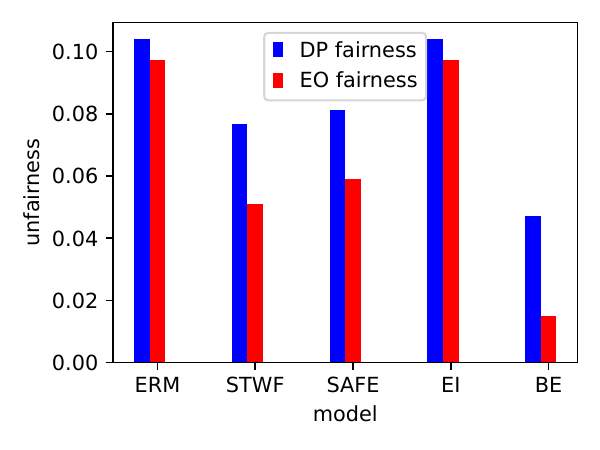}
\includegraphics[width=0.28\textwidth]{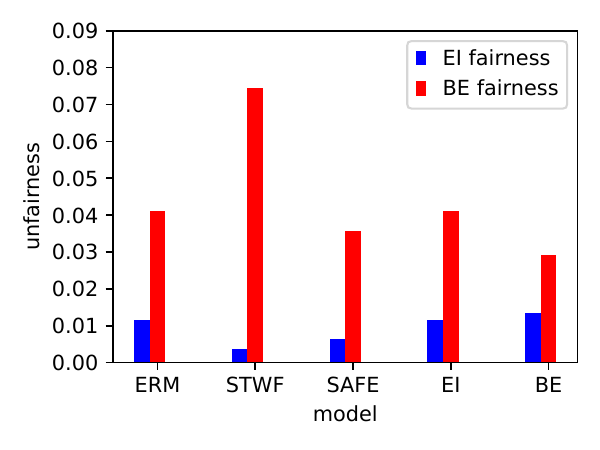}
\caption{Comparisons under the ACSincome-CA dataset.}
\label{subfig:income_comparison}
\end{subfigure}
\caption{Comparisons of welfare (left plot) and unfairness (middle \& right plot) for different algorithms.}
\label{fig:comparison_main}
\end{figure*}

We conduct comprehensive experiments on one synthetic and two real-world datasets to validate \texttt{STWF}\footnote{\url{https://github.com/osu-srml/StrategicWelfare}}. We first illustrate the trade-offs for each pair of welfare and examine how they can be balanced by adjusting $\lambda_1, \lambda_2$ in Alg.~\ref{alg:welfare}. Then, we compare \texttt{STWF} with previous algorithms that only consider the welfare of a subset of parties. Since many of the existing algorithms also consider fairness across agents from different groups, we also evaluate the unfairness induced by our algorithm in addition to decision-maker, agent, and social welfare. We use a three-layer ReLU neural network to learn labeling function $h$ and logistic regression to train the policy $f$, and $D(x) = \mathbf{1}(f(x) \ge 0.5)$. For each parameter setting, we perform 20 independent runs of the experiments with different random seeds and report the average.

\subsection{Datasets} Next, we introduce the data used in the experiments. For all datasets, $80\%$ of the data is used for training.

\begin{itemize}[leftmargin=0.5cm]
    \item 
\textbf{Synthetic data.} It has two features $X = (X_1, X_2)$ and a  sensitive attribute $Z \in \{0,1\}$ for demographic groups. Let  $(X|Z = z)$ be multivariate Gaussian random variables with different means and variances conditioned on $Z$. Labeling function $h(x)$ is a quadratic function in both $x_1, x_2$, and the label $Y = \mathbf{1}(h(X) \ge 0.5)$. We generate $10000$ independent samples with $20\%$ of them having $Z = 1$ and the rest $Z = 0$. We assume all agents have the same cost function $c(x,x') = 5 \cdot \|x'-x\|^2$.

\item \textbf{ACSIncome-CA data} \citep{ding2021retiring}. It contains income-related data for California residents. Let label $Y$ be whether an agent's annual income is higher than $50$K USD and let the sensitive attribute $Z$ be \texttt{sex}. We randomly select $20000$  samples from the data of year $2018$. We assume all agents have the same cost function $c(x,x') = \|x'-x\|^2$. 

\item \textbf{German Credit data} \citep{Dua2019}. It has $1000$ samples and the label $Y$ is whether the agent's credit risk is high or low. We adopt the data preprocessed by \cite{jiang2020identifying} where \texttt{age} serves as the sensitive attribute to split agents into two groups. We assume the agents are able to improve the following features: \texttt{existing account status}, \texttt{credit history}, \texttt{credit amount}, \texttt{saving account}, \texttt{present employment}, \texttt{installment rate}, \texttt{guarantors}, \texttt{residence}. 
We assume all agents have the same cost function $c(x,x') = \|x'-x\|^2$. 

\end{itemize}

\subsection{Balance the Welfare with Different $\lambda_1, \lambda_2$}\label{subsec:wdynamics}

We first validate \texttt{STWF} and show that adjusting the hyper-parameters $\lambda_1, \lambda_2$ effectively balances the welfare of different parties. The results in Fig.~\ref{fig:adjust_single} are as expected, where \textit{social welfare} $\text{SWF}(f)$ increases in $\lambda_1$ and \textit{agent welfare} $\text{AW}(f)$ increases in $\lambda_2$. This shows the effectiveness of using our algorithm to adjust the allocation of welfare under different situations. As we analyzed in Sec.~\ref{sec:welfare}, under the general non-linear setting, each pair of welfare possibly conflicts. Fig.~\ref{fig:adjust_single} validates this by demonstrating the trade-offs between different welfare pairs on two datasets, while it also reveals that the conflicts are not always serious between each pair. For the synthetic data, increasing \textit{social welfare} slightly conflicts with \textit{decision-maker welfare}, while increasing \textit{agent welfare} drastically conflicts with the \textit{decision-maker welfare}. But the trade-off between \textit{social welfare} and \textit{agent welfare} is not obvious. For the German Credit dataset, increasing \textit{social welfare} seems to slightly conflict with both \textit{agent welfare} and \textit{decision-maker welfare}, while increasing \textit{agent welfare} has an obvious trade-off with \textit{decision-maker} welfare. Conversely, for the ACSIncome-CA dataset, increasing \textit{social welfare} only sacrifices \textit{agent welfare}, but increasing \textit{agent welfare} incurs conflicts with both other welfare. 

\subsection{Comparison with Previous Algorithms}\label{subsec:wfairness}

Since \texttt{STWF}  is the first optimization protocol that simultaneously considers the welfare of the decision-maker, agent, and society, there exists no algorithm that we can directly compare with. Nonetheless, we can adapt existing algorithms that only consider the welfare of a subset of parties to our setting, and compare them with \texttt{STWF}.


Specifically, we compare \texttt{STWF} with four algorithms: 

\begin{itemize}[leftmargin=0.5cm]
  \item \textbf{Empirical Risk Minimization (\texttt{ERM}):} It only considers \textit{decision-maker welfare} by minimizing the predictive loss on agent current data.   \item \textbf{Safety (\texttt{SAFE})} \citep{Rose2020}: It considers both \textit{decision-maker welfare} and \textit{agent safety}. The goal is to train an accurate model without hurting agent qualification.    \item\textbf{Equal Improvement (\texttt{EI})} \citep{guldogan2022equal}: It considers \textit{agent improvement} by equalizing the probability of unqualified agents becoming qualified from different groups, as opposed to maximizing agent improvement considered in our work. 
  \item \textbf{Bounded Effort (\texttt{BE})}  \citep{heidari2019long}: It considers \textit{agent improvement} by equalizing the proportion of new agents becoming qualified after the best response. The difference between $\texttt{BE}$ and $\texttt{EI}$ is a bit subtle, where we show details in Tab. \ref{tab:fairdef}.
\end{itemize}

Note that both \texttt{EI} and \texttt{BE} focus on fairness across different groups. To comprehensively compare different algorithms, we also evaluate the unfairness induced by each algorithm. Specifically, we consider both fairness notions with and without considering agent behavior, including Equal Improvability (\textsf{EI}), Bounded Effort (\textsf{BE}), Demographic Parity (\textsf{DP}), and Equal Opportunity (\textsf{EO}), which are defined in Tab.~\ref{tab:fairdef}. We run all experiments on a MacBook Pro with Apple M1 Pro chips, memory of 16GB, and Python 3.9.13. In all experiments of Sec.~\ref{subsec:wfairness}, we use the \texttt{ADAM} optimizer. Except \texttt{ERM}, we first perform hyper-parameter selections for all algorithms, e.g., we select $\lambda_1, \lambda_2$ of \texttt{STWF} by grid searching with cross-validation. For \texttt{STWF}, we perform cross-validation on 7 different seeds to select the learning rate from $\{0.001,0.01,0.1\}$ and $\lambda_1, \lambda_2 \in \{0,0.5,1,1.5,2\}$. For \texttt{EI}, \texttt{BE} and \texttt{SAFE}, we choose the hyperparameters in the same way as the previous works \cite{guldogan2022equal,Rose2020,heidari2019long}. Specifically, $\lambda$ (the strength of the regularization) is $0.1$ for all these algorithms.

\begin{table*}[t]
    \caption{Fairness notions used in our experiments.}
    \vspace{-0.2cm}\label{tab:fairdef}
    \centering
    \resizebox{0.7\textwidth}{!}{
    \centering
    \begin{tabular}{lc}
        \toprule
        Notion & Definition  \\
        \toprule
       Equal Improvability  (\textsf{EI}) & $ P\left( f(x^{*})\hspace{-0.5mm} \ge \hspace{-0.5mm} 0.5 \mid f(x)  \hspace{-0.5mm} <  \hspace{-0.5mm} 0.5, Z = z \hspace{-0.5mm} \right) \hspace{-0.5mm} = \hspace{-0.5mm} P\left(f(x^{*}) \hspace{-0.5mm} \ge \hspace{-0.5mm} 0.5 \mid f(x) \hspace{-0.5mm} < \hspace{-0.5mm} 0.5 \hspace{-0.5mm} \right)$  \\
       Bounded Effort (\textsf{BE})  & $P\left(f(x^{*}) \ge 0.5, f(x) < 0.5 \mid Z = z\right) = P\left(f(x^{*}) \ge 0.5, f(x) < 0.5 \right)$\\
       Demographic Parity (\textsf{DP}) & $ P\left(f(x) \ge 0.5\mid Z=z\right) = P\left(f(x) \ge 0.5\right)$ \\
        Equal Opportunity (\textsf{EO}) & $P\left(f(x) \ge 0.5\mid Y=1,Z=z\right) = P\left(f(x) \ge 0.5\mid Y=1\right)$  \\
        \bottomrule
    \end{tabular}
    }
\end{table*}

\subsubsection{Comparison of welfare.} Tab.~\ref{results-table:comparison_main} summarizes the performances of all algorithms, where \texttt{STWF} produces the largest total welfare on the synthetic dataset and the ACSIncome-CA dataset. For the German Credit dataset, all algorithms produce similar welfare, while the \texttt{ERM} has a slight advantage. More clearly, as shown in the left plot from Fig.~\ref{subfig:quad_comparison} to Fig.~\ref{subfig:income_comparison}, \texttt{STWF} gains a large advantage, demonstrating the potential benefits of training ML model in a welfare-aware manner. Meanwhile, although \texttt{STWF} does not have a similar advantage on the German Credit dataset, it produces a much more balanced welfare allocation, suggesting the ability of \texttt{STWF} to take care of each party under the strategic learning setting. However, since we select $\lambda_1, \lambda_2$ only based on the sum of the welfare, it is possible to sacrifice the welfare of some specific parties (e.g., \textit{agent welfare} in the ACSincome-CA dataset). This can be solved by further adjusting $\lambda_1, \lambda_2$ based on other selection criteria.

\begin{figure}[h]
\centering
\includegraphics[width=0.25\textwidth]{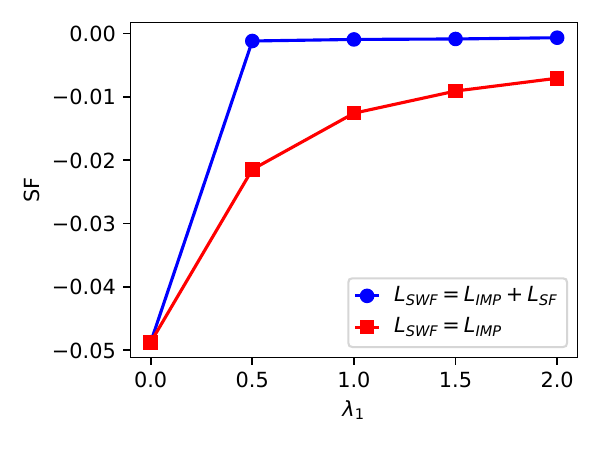}
\vspace{-0.3cm}
\caption{Comparisons of safety under the synthetic dataset.}
\label{fig:imp_safe_german}
\end{figure}

Besides, we also illustrate that safety is not automatically guaranteed if we only aim to maximize agent improvement in the synthetic dataset. Specifically, if we let $L_{SWF} = L_{IMP}$, we can visualize the safety in Fig.~\ref{fig:imp_safe_german}. It is obvious that only when $L_{SF}$ is added into $L_{SWF}$ will the decision rule become perfectly safe as $\lambda_1$ increases.

\subsubsection{Improve welfare does not ensure fairness.} The middle and the right plots of Fig.~\ref{fig:comparison_main} compare the unfairness of different algorithms. Although \texttt{STWF} has the highest welfare, it does not ensure fairness. These results further shed light on a thought-provoking question: how likely is it for a fairness-aware algorithm to ``harm" each party's welfare under the strategic setting? Recent literature on fairness \citep{guldogan2022equal,heidari2019long} already began to consider ``agent improvement" when strategic behaviors are present. However, it remains an intriguing direction to consider fairness and welfare together.

\section{Conclusion \& Societal Impacts}

To facilitate socially responsible decision-making on strategic human agents in practice, our work is the first to consider the welfare of all parties in strategic learning under non-linear settings when the agents have specific information levels. Our algorithm relies on the accurate estimation of agent response. In practice, the agent's behavior can be highly complex. With incorrect estimations of agents' behavior, our algorithm may fail and lead to unexpected outcomes. Thus, it is worthwhile for future work to collect real data from human feedback, and learn agent response function. Additionally, our experiments demonstrate that welfare and fairness may not imply each other, making it necessary to consider both notions theoretically to further promote trustworthy machine learning.

\section*{Acknowledgement}

This material is based upon work supported by the U.S. National Science Foundation under award IIS-2202699, by grants from the Ohio State University's Translational Data Analytics Institute and College of Engineering Strategic Research Initiative.

\bibliography{aaai24}

\appendix

\appendix

\onecolumn

\section{Notes}

Compared with the conference version, this draft clarifies that the cost functions in the welfare analysis are quadratic, which was previously an implicit assumption. We also add $2$ missing conditions in Theorem~\ref{thm:general}, namely that $\mathcal{F}$ must be linear with weights on a ellipsoidal surface parameterized by $R$, and the weight vector of $h$ must be the eigenvector of $R^{-1}Q^{-1}$; otherwise, $\text{IMP}(f)$ is not guaranteed to be maximized when $f = h$. These clarifications do not affect the main claims or the proposed algorithms. In fact, they highlight that aligning welfare across all parties is even more challenging under these conditions, thereby motivating STWF even in a stronger way.

\section{Proofs}\label{app:proof}

\subsection{Proof of Prop. \ref{prop:info}}\label{app:infopf}

\begin{proof}

(i) Proof of the necessity: The fact that agents are guaranteed to best respond means they can estimate $f$ perfectly. Since the agents have information level $K$, they will estimate $f(x)$ by the Taylor expansion of $f(x)$ at $x$ using gradients up to the $K^{th}$ order and this results in a polynomial function with order at most $K$ as specified in Eqn. \eqref{eq:taylor}. Note that if the gradient of any order does not exist, the Taylor expansion will definitely not yield a correct estimation of $f$. 

\begin{align}\label{eq:taylor}
    f(x') = &f(x) + \sum_{i=1}^{d} \frac{\partial f}{\partial x_i}(x) (x'_i - x_i) + \frac{1}{2!} \sum_{i=1}^{d} \sum_{j=1}^{d} \frac{\partial^2 f}{\partial x_i \partial x_j}(x) (x'_i - x_i)(x'_j - x_j) + \cdots \\
    \nonumber &+ \frac{1}{K!} \sum_{i_1=1}^{d} \cdots \sum_{i_K=1}^{d} \frac{\partial^K f}{\partial x_{i_1} \cdots \partial x_{i_K}}(x) (x'_{i_1} - x_{i_1})\cdots(x'_{i_K} - x_{i_K})
\end{align}

(ii) Proof of sufficiency: When $f$ is a polynomial function with order $k \ge K$, the agents with information level $K$ are guaranteed to know all levels of gradients of $f$ which are non-zero. Then the agents can construct $f$ perfectly according to Eqn. \eqref{eq:taylor}, then the agent response according to Def. \ref{def:br} is indeed the best response.
\end{proof}

\subsubsection{Proof of Thm. \ref{thm:general}}\label{app:generalpf}

\begin{proof}

1. Proof of necessity: Since the maximization of the decision-maker welfare regardless of $P_X$ needs $f = h$ (otherwise, one can always take the point mass on the point that $f$ misidentifies as the new feature distribution and $f$ is no longer the one maximizing the \textit{decision-maker welfare}), then $h \in \mathcal{F}$ is needed. Then the agents need to best respond to $f$ to ensure the safety. If there is an arbitrary agent who does not best respond to $f = h$, this means she misunderstands $h$ and this can result in deteriorating qualification. Thus, according to Prop. \ref{prop:info}, $h$ itself must be a polynomial of $X$ with at most order $K$ to guarantee agents to best respond. 

Next, consider the situation when $\mathcal{F}$ consists not only linear functions with normalized weights: (i) Firstly, if $\mathcal{F}$ is not linear, we can construct a counter-example: consider $x \in [0,1]$ in $1$-d space and $h(x) = x$. $Q = 0.6$, and the population is a point mass at $0$. Then if we deploy $f(x) = x^2$, then every agent with $x = 0$ will change their features to $x = 1$, but deploying $f(x) = x$ will only result in their changing to $x = 0.833$. Therefore, $f = h$ does not result in largest $\text{IMP}(f)$. This example demonstrates that we cannot even relax $\mathcal{F}$ to the quadratic function class; (ii) Next, if $\mathcal{F}$ is linear and $h = \hat{w}^Tx$, $f = w^Tx$, we first solve the best response as follows:

Let $d:=x'-x$. We need to derive $\max_{d\in\mathbb{R}^d}\ \ w^\top d - d^\top Q d$ and this gives $\nabla_d\big(w^\top d - d^\top Q d\big)= w - 2Qd = 0 \Longrightarrow\  d^{*}(w) = \frac{1}{2}Q^{-1}w$. Thus, $x'(w) = x + d^{*}(w) = x + \frac{1}{2}Q^{-1}w$. Next, $\text{IMP}(f) = \frac{1}{2}(Q^{-1}\hat{w})^Tw$. Obviously, when $w$ is unbounded, $\text{IMP}(f)$ can be infinite. To always let $\hat{w}$ to be the maximizer of itself, $\hat{w}$ must belong to an ellipsoidal surface (e.g., a ball surface). Assume the surface can be written as $\{w | w^\top Rw = c^2\}$, then we can solve the maximizer $w^{*} = c\frac{R^{-1}Q^{-1}\hat{w}}{\sqrt{(Q^{-1}\hat{w})^\top R^{-1}(Q^{-1}\hat{w})}}$. To let $w^{*} = \hat{w}$, we need $Q^{-1}\hat{w}$ and $R\hat{w}$ have the same direction, which is equivalent to $\hat{w}$ is the eigenvector of $R^{-1}Q^{-1}$.

2. Proof of sufficiency: When $h$ is linear and $\mathcal{F}$ is also linear, the decision-maker can directly deploy $h = f$ to ensure the maximization of $\text{DW}(f), \text{SF}(f), \text{AW}(f)$. Next, according to the above derivation, $\text{IMP}(f)$ is also maximized.
    
\end{proof}

\subsubsection{Proof of Prop. \ref{prop:prop2}}\label{app:prop2}

\begin{proof}
1. Proof of the necessity: Since the maximization of the decision-maker welfare needs $f = h$, then $h \in \mathcal{F}$ is needed. Then the agents need to best respond to $f$ to ensure the maximization of social welfare. If there is an arbitrary agent who does not best respond to $f = h$, this means she misses the largest possible improvement and the overall maximization of social welfare is failed. Thus, according to Prop. \ref{prop:info}, $h$ itself must be a polynomial of $X$ with at most order $K$ to guarantee agents to best respond. Finally, to ensure the maximization of $\text{IMP}(f)$, $\mathcal{F}$ must be linear.

2. Proof of sufficiency: Thm. \ref{thm:general} is already a sufficient condition to guarantee the compatibility of \textit{decision-maker welfare} and \textit{social welfare}.
\end{proof}

\subsection{Proof of Thm.~\ref{theorem:generalswf}}\label{app:swfproof}
\begin{proof}
1. Proof of the necessity: If the condition does not hold this means there must exist a cost function $c(x,x')$, under which we can find a region $A \subset \text{dom}(X)$ where $Q^{f^{*}}_x$ increases while $h$ decreases inside $A$. Thus, if we specify a cost function and an agent to satisfy the requirement that the agent can only change her features within $A$ and when $P_X$ is any continuous distribution, then $f^{*}$ is unsafe.

2. Proof of the sufficiency: We know for each dimension $i$, $Q^{f^{*}}_x$ improves at the same direction as $h$ and only the magnitude differs. This means agents will never be misguided by $Q^{f^{*}}_x$, thereby ensuring the safety.
\end{proof}

\subsection{Proof of Thm.~\ref{theorem:generalaw}}\label{app:awproof}
\begin{proof}
1. Proof of the necessity: When (i) does not hold, $f^{*}$ is not equal to $h$ and the agents cannot perfectly estimate $f^{*}$ with $Q^{f^{*}}_x$. Therefore, if (ii) also does not hold, this means any classifier $f'$ maximizing the \textit{agent welfare} does not let the agents get an estimation $Q^{f'}_{x}$ only differing in constant to $Q_{imp, x}$ which is the estimation for $f^{*}$. Thus, we can find a region $A \subset \text{dom}(X)$ where $Q^{f^{*}}_x$ where the $\max_A Q^{f^{*}}_x - \min_A Q^{f^{*}}_x$ does not equal to $\max_A Q^{f'}_{x} - \min_A Q^{f'}_{x}$. Then if we find an agent and a cost function to let the agent can only modify her features within $A$, her improvement under $f'$ will be different from her improvement under $f^{*}$, which means $f'$ cannot maximize \textit{social welfare}.

2. Proof of the sufficiency: (i) is already a sufficient condition. Then for (ii), we know $Q^{f'}_{x}$ induces the same amount of improvement as $Q^{f^{*}}_x$, which completes the proof.

\end{proof}

\section{Plots with error bars}\label{app:ebar}

\begin{figure}[h]
\centering
\begin{subfigure}[b]{0.9\textwidth}
\centering
\includegraphics[width=0.3\textwidth]{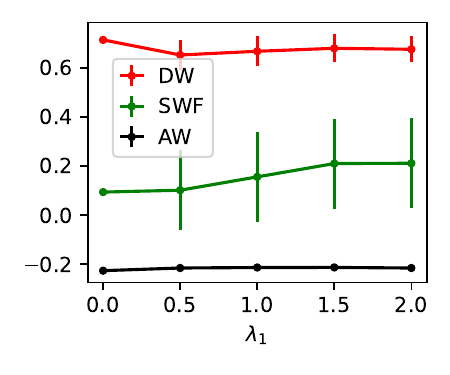}
\includegraphics[width=0.3\textwidth]{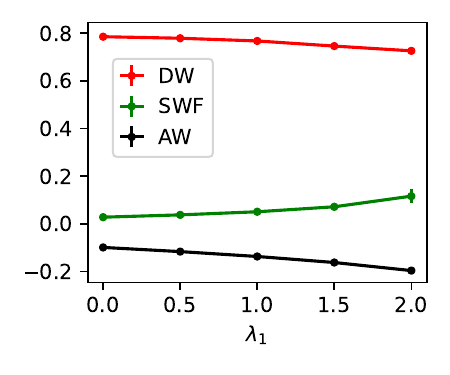}
\includegraphics[width=0.3\textwidth]{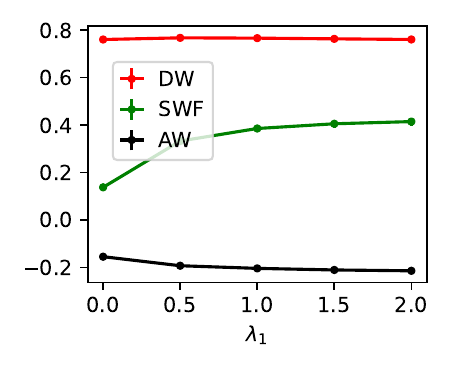}
\vspace{-0.2cm}
\caption{Dynamics of welfare while increasing $\lambda_1$ and keeping $\lambda_2 = 0$ in Eqn.\eqref{eq:opt}.}
\label{subfig:ebar_lambda1}
\end{subfigure}
\hfill
\begin{subfigure}[b]{0.9\textwidth}
\centering
\includegraphics[width=0.3\textwidth]{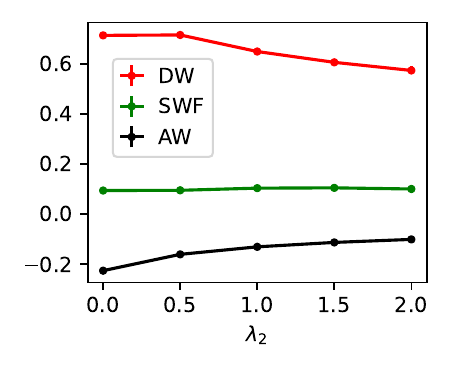}
\includegraphics[width=0.3\textwidth]{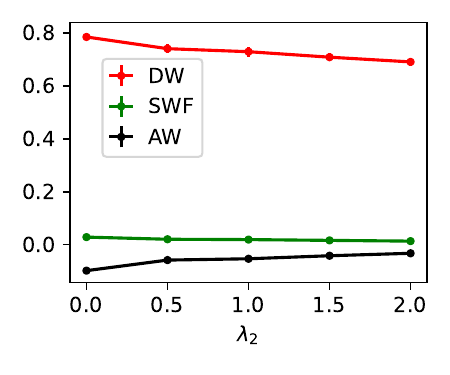}
\includegraphics[width=0.3\textwidth]{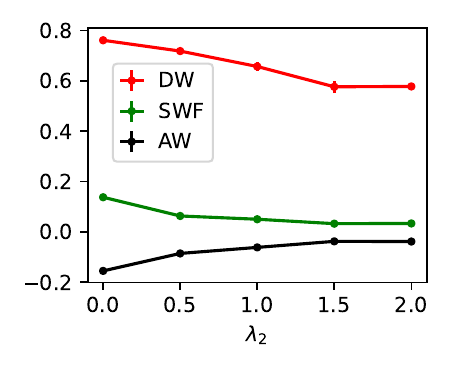}
\vspace{-0.2cm}
\caption{Dynamics of welfare while increasing $\lambda_2$ and keeping $\lambda_1 = 0$ in Eqn.\eqref{eq:opt}.}
\label{subfig:ebar_lambda2}
\end{subfigure}
\caption{Error bar version of Fig.~\ref{fig:adjust_single}}
\label{fig:adjust_single_ebar}
\vspace{-0.2cm}
\end{figure}

\begin{figure}
\centering
\begin{subfigure}[b]{0.9\textwidth}
\centering
\includegraphics[width=0.3\textwidth]{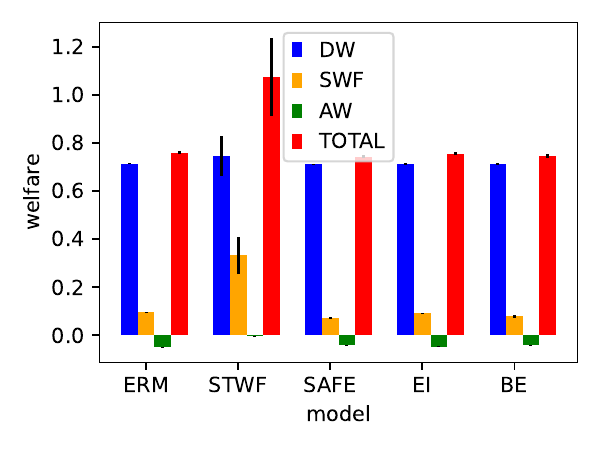}
\includegraphics[width=0.3\textwidth]{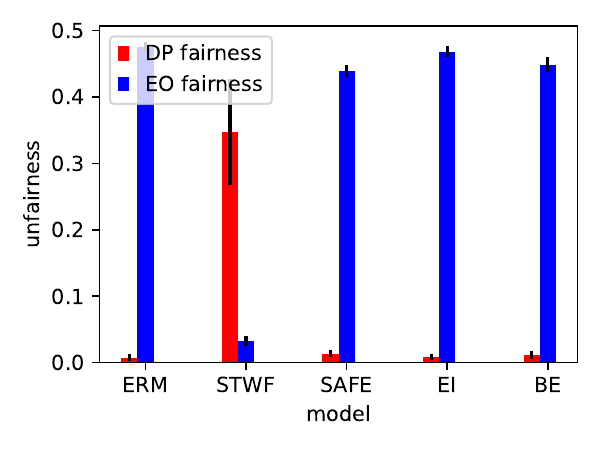}
\includegraphics[width=0.3\textwidth]{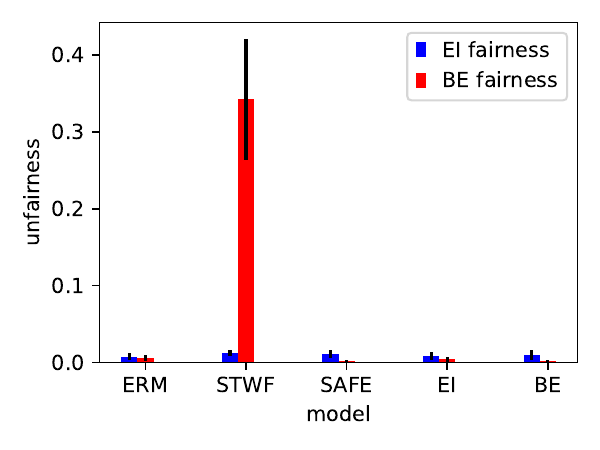}
\vspace{-0.2cm}
\caption{Comparisons under the synthetic dataset.}
\label{subfig:ebar_quad_comparison}
\end{subfigure}
\hfill
\begin{subfigure}[b]{0.9\textwidth}
\centering
\includegraphics[width=0.3\textwidth]{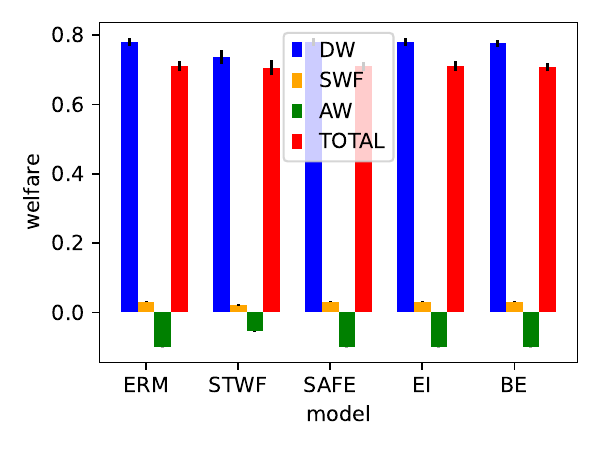}
\includegraphics[width=0.3\textwidth]{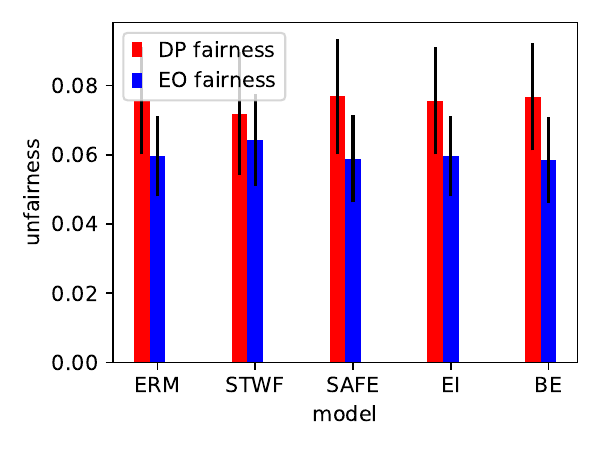}
\includegraphics[width=0.3\textwidth]{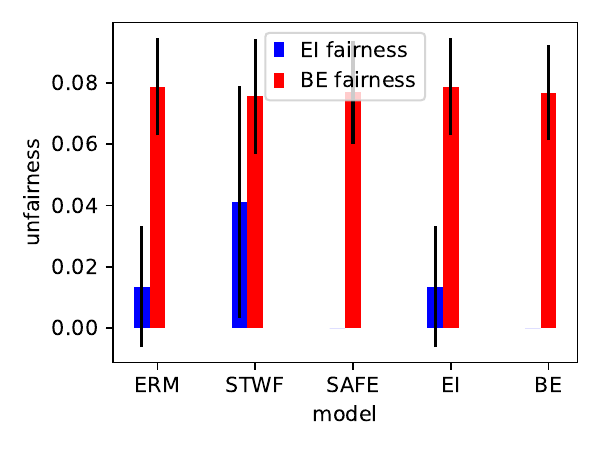}
\vspace{-0.2cm}
\caption{Comparisons under the German Credit dataset.}
\label{subfig:ebar_german_comparison}
\end{subfigure}
\hfill
\begin{subfigure}[b]{0.9\textwidth}
\centering
\includegraphics[width=0.3\textwidth]{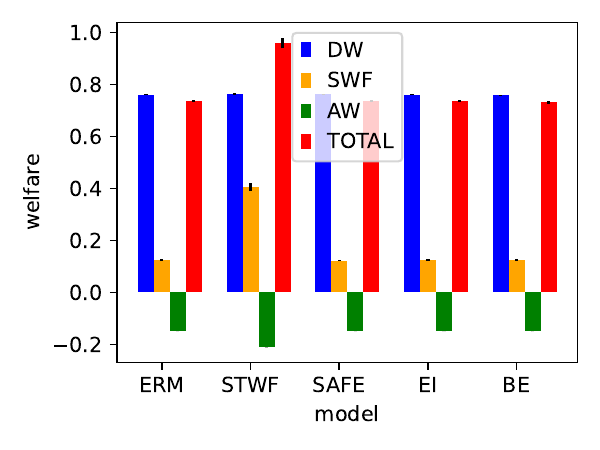}
\includegraphics[width=0.3\textwidth]{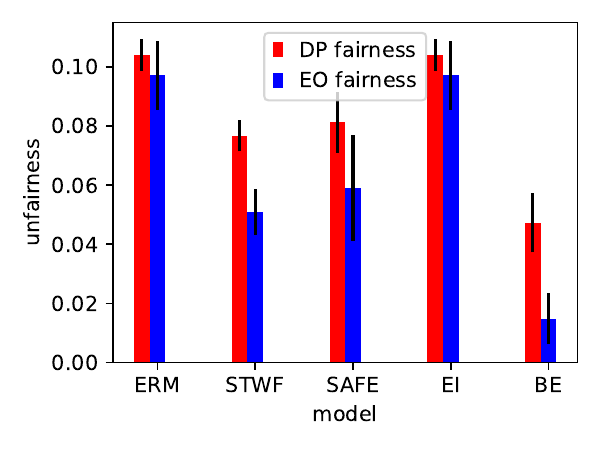}
\includegraphics[width=0.3\textwidth]{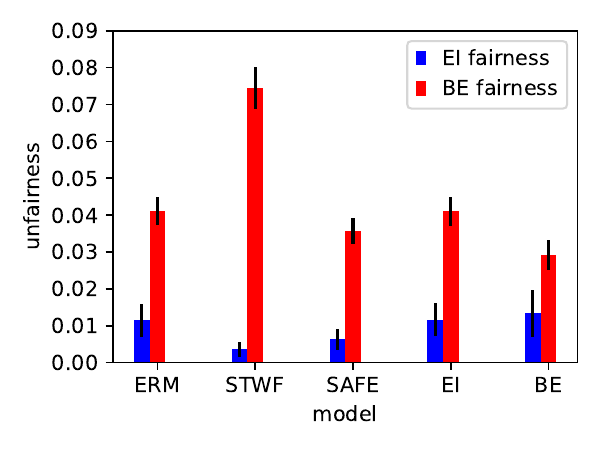}
\vspace{-0.2cm}
\caption{Comparisons under the ACSincome-CA dataset.}
\label{subfig:ebar_income_comparison}
\end{subfigure}
\caption{Error bar version of Fig.~\ref{fig:comparison_main}.}
\label{fig:comparison_ebar}
\vspace{-0.2cm}
\end{figure}

We provide the error bar version of the previous figures as follows (Fig.~\ref{fig:adjust_single_ebar}, Fig.~\ref{fig:comparison_ebar}).

\end{document}